\documentclass[10pt,twocolumn,letterpaper]{article}

\usepackage[pagenumbers]{iccv} 

\usepackage{graphicx}
\usepackage{diagbox}

\usepackage{pgfplotstable}
\usepackage{pgfplots}
\usepackage{booktabs}
\usepackage{colortbl}

\usepackage{subcaption}
\usepackage{arydshln}
\usepackage{multirow}
\usepackage{tikz}
\usepackage{pgfplots}
\pgfplotsset{compat=1.18}

\usepackage{comment}

\definecolor{iccvblue}{rgb}{0.21,0.49,0.74}
\usepackage[pagebackref,breaklinks,colorlinks,allcolors=iccvblue]{hyperref}

\def\confName{ICCV}
\def\confYear{2025}

\def\l1{\ensuremath{\ell_1}\xspace}
\def\l2{\ensuremath{\ell_2}\xspace}

\makeatletter
\DeclareRobustCommand\onedot{\futurelet\@let@token\@onedot}
\def\@onedot{\ifx\@let@token.\else.\null\fi\xspace}

\def\ie{\emph{i.e}\onedot}

\makeatother

\newcommand{\Lcls}{\ensuremath{\mathcal{L}_{\mbox{domain}}}}
\newcommand{\Leb}{\ensuremath{\mathcal{L}_{\mbox{pretr.embed.}}}}
\newcommand{\Ltheta}{\ensuremath{\mathcal{L}_{\mbox{pretr.params}}}}
\newcommand{\lamtheta}{\ensuremath{\lambda_\theta}}
\newcommand{\lameb}{\ensuremath{\lambda_{\text{emb}}}}

\definecolor{tolblue}{rgb}{0,0.447,0.702}
\definecolor{tolorange}{rgb}{0.902,0.624,0}
\definecolor{tolgreen}{rgb}{0,0.62,0.45}
\definecolor{tolred}{rgb}{0.835,0.369,0}

\newcommand\blfootnote[1]{%
  \begingroup
  \renewcommand\thefootnote{}\footnote{#1}%
  \addtocounter{footnote}{-1}%
  \endgroup
}

\title{Infusing fine-grained visual knowledge to Vision-Language Models}

\author{%
Nikolaos-Antonios Ypsilantis$^{*1}$ \quad Kaifeng Chen$^{2}$ \quad André Araujo$^2$ \quad Ond\v{r}ej Chum$^1$ \and \\
$^1$VRG, FEE, Czech Technical University in Prague \quad $^2$Google DeepMind
}

\begin{document}
\maketitle
\begin{abstract}
    Large-scale contrastive pre-training produces powerful Vision-and-Language Models (VLMs) capable of generating representations (embeddings) effective for a wide variety of visual and multimodal tasks. However, these pretrained embeddings remain suboptimal for fine-grained open-set visual retrieval, where state-of-the-art results require fine-tuning the vision encoder using annotated domain-specific samples. Naively performing such fine-tuning typically leads to catastrophic forgetting, severely diminishing the model’s general-purpose visual and cross-modal capabilities.
    
    In this work, we propose a fine-tuning method explicitly designed to achieve optimal balance between fine-grained domain adaptation and retention of the pretrained VLM’s broad multimodal knowledge. Drawing inspiration from continual learning literature, we systematically analyze standard regularization techniques aimed at knowledge retention and propose an efficient and effective combination strategy. Additionally, we address the commonly overlooked yet critical aspects of validation set design and hyperparameter tuning to ensure reproducibility and robust generalization across datasets and pretrained models. We extensively evaluate our method on both fine-grained and coarse-grained image-image and image-text retrieval benchmarks. Our approach consistently achieves strong results, notably retaining the visual-text alignment without utilizing any text data or the original text encoder during fine-tuning. Code and model checkpoints: \texttt{https://github.com/nikosips/infusing}.
    
    \blfootnote{$^*$Corresponding author: \texttt{ypsilnik@fel.cvut.cz}}
    
\end{abstract}

\section{Introduction}
\begin{figure}[htbp]
\centering
\hspace{-10pt}
\scalebox{0.93}{
\begin{tikzpicture}
\begin{axis}[
    ybar,
    bar width=0.8cm,
    width=9cm,
    height=8cm,
    ylabel={$\Delta$Performance},
    ylabel style={font=\large},
    xtick={1, 2, 3},
    xticklabels={In-Domain \\ Image-Image, Out-Of-Domain \\ Image-Image, Out-Of-Domain \\ Image-Text},
    xticklabel style={align=center, rotate=0},
    xmin=0.5,
    xmax=3.5,
    axis line style={thick},
    tick style={draw=none},
    legend style={at={(0.7,0.95)}, anchor=north, legend columns=1},
    nodes near coords, 
    point meta=y,         
    every node near coord/.append style={
        font=\small
    },
]

\addplot+[ybar,
          bar shift=-0.4cm,
          fill=cyan!100, draw=black,
          nodes near coords,
          every node near coord/.append style={
            text=cyan!100,
            font=\scriptsize,
            /pgf/number format/showpos,
          }
] coordinates {
  (1,+31) (2,-6) (3,-12.7)
};
\addlegendentry{Standard fine-tuning} 

\addplot+[ybar,
          bar shift=+0.4cm,
          fill=orange!100, draw=black,
          nodes near coords,
          every node near coord/.append style={
            text=orange!100,
            font=\scriptsize,
            /pgf/number format/showpos,
          }
] coordinates {
  (1,+29.9) (2,-0.1) (3,-0.9)
};
\addlegendentry{Our fine-tuning}

\draw[black, thick] 
  (axis cs:-2,0) -- (axis cs:10,0);

\end{axis}
\end{tikzpicture}
}

\caption{Performance change ($\Delta$) between the pretrained model to the fine-tuned model (SigLIP ViT-Base/16, SOP fine-tuning).
Our fine-tuning recipe effectively mitigates out-of-domain knowledge forgetting and retains the visual alignment with the textual encoder of the VLM, without sacrificing in-domain specialization.
}
\label{fig:accuracy_comparison}

\vspace{-6pt}

\end{figure}
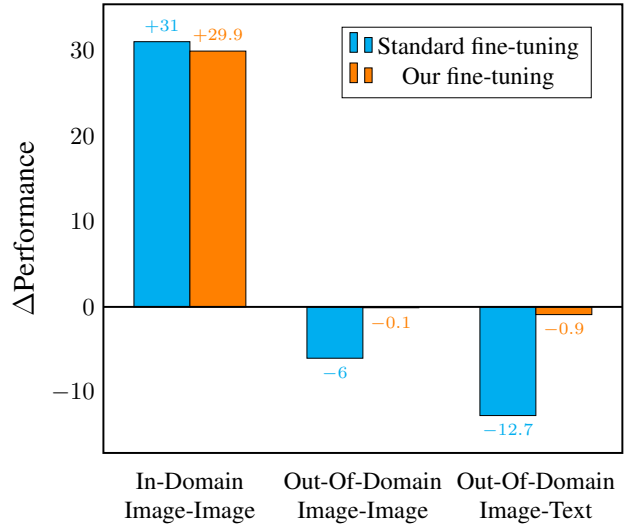
\label{sec:intro}

Vision-and-Language Models (VLMs), such as CLIP~\cite{radford2021learning}, SigLIP~\cite{zhai2023sigmoid}, and TIPS~\cite{maninis2025tips}, have demonstrated remarkable performance across diverse visual and multimodal tasks through large-scale pretraining on extensive collections of image-caption pairs. These models learn rich, general-purpose multimodal embeddings that enable effective cross-modal and visual retrieval via embedding-based similarity search. Consequently, VLMs have become foundational tools across numerous applications, ranging from general-purpose image retrieval to specialized domains.

One critical specialized application area is fine-grained open-set visual retrieval. Such tasks include clothing retrieval~\cite{song2016deep}, food recognition~\cite{min2023large}, and species identification~\cite{van2018inaturalist}, to name just a few, where models must discriminate among visually similar yet semantically distinct categories. 
In these cases, only the vision encoder of the VLM is typically utilized, employing it to index large-scale image databases offline, with image-based queries processed at inference time.
In order to be effective, the vision encoder is trained using expert-labeled fine-grained datasets, yet deployed to perform retrieval on unseen classes; hence the term \emph{open-set}. 

While pretrained VLM vision encoders represent the current state-of-the-art for general-purpose visual understanding~\cite{zhai2023sigmoid,maninis2025tips,tschannen2025siglip,ksm+25}, they do not achieve optimal performance on fine-grained recognition tasks without domain-specific fine-tuning~\cite{ycc+23}. 
Current best practices~\cite{ermolov2022hyperbolic,ycc+23,ptm22,ramzi2025optimization,venkataramanan2022takes,ypsilantis2024udon} involve fine-tuning the vision encoder with domain-specific labeled data, significantly boosting performance within the target domain. 
However, this approach carries significant limitations. First, fine-tuning on domain-specific data typically weakens the model’s general-purpose capabilities, diminishing performance on out-of-domain visual tasks and reducing cross-modal alignment (see Fig.~\ref{fig:accuracy_comparison}). Second, the deterioration of generalizable capabilities compels practitioners to maintain multiple specialized models for different tasks, increasing computational overhead, storage requirements, and operational complexity.

In this paper, we propose a fine-tuning strategy specifically designed to address these challenges. 
Our primary objective is to ``infuse'' fine-grained visual knowledge to a VLM, \ie enhance a VLM's fine-grained recognition capabilities on a specific visual domain without sacrificing its original general-purpose visual understanding and cross-modal retrieval performance. 
Inspired by techniques from continual learning, we adopt two complementary regularization approaches: parameter-space regularization towards the pretrained model and embedding-space regularization via pretrained representation distillation. While these techniques individually have precedents in the continual learning literature, our work explores their integration and behavior specifically within the fine-grained visual retrieval context. We provide empirical insights on effectively combining these regularization techniques, achieving a robust fine-tuning framework that maintains a strong balance between fine-grained domain specialization and broader multimodal representation quality.

Moreover, we address methodological shortcomings in the existing literature, notably the neglect of validation strategies essential for accurately predicting both in-domain and out-of-domain generalization performance during fine-tuning. We propose a rigorous validation framework, leveraging specific validation sets to transparently and reliably evaluate these dual performance aspects. Our validation methodology ensures robust hyperparameter tuning, promoting the broader applicability and reproducibility of our approach.

Our contributions are summarized as follows:

\begin{itemize}
    \item We perform an analysis of two different types of regularization losses that are widely used in continual learning, investigating how findings for each one of them translate to the fine-grained open set visual retrieval setting when tasked to retain the knowledge of the pretrained model.
    \item We propose a fine-tuning method for preserving general-purpose multimodal capabilities while significantly enhancing fine-grained open-set visual retrieval performance in VLMs, addressing an important yet unexplored problem (see Fig.~\ref{fig:accuracy_comparison}).
    \item We offer a transparent validation set design, explicitly aimed at reliably predicting both in-domain specialization and out-of-domain generalization.
    \item We conduct extensive empirical evaluations on large-scale, realistic datasets to demonstrate the effectiveness and robustness of our method for both image-only and image-text retrieval tasks, highlighting its practical value and generalizability.
\end{itemize}
\section{Related Work}
\label{sec:related_work}

We describe prior art that is related to our work, split into different topics.
\vspace{-10pt}
\paragraph{Fine-Tuning Pretrained Vision Encoders without Catastrophic Forgetting.}
The standard approach for adapting pre-trained vision encoders, such as the Vision Transformers (ViTs) of Vision-Language Models (VLMs), to downstream tasks involves full fine-tuning.
Full fine-tuning typically yields the highest in-domain performance, particularly when combined with metric learning objectives for retrieval or fine-grained recognition tasks~\cite{kumar2022fine,ptm22,ramzi2025optimization,he2022transfg,suma2023large,ypsilantis2024udon}.
However, this often leads to \textit{catastrophic forgetting}: the erosion of general-purpose visual features learned during pretraining, as the model over-specializes to the fine-tuning task~\cite{mukhoti2024ldifs,kumar2022fine}.

One way to mitigate forgetting is partial fine-tuning strategies. 
These approaches address the problem by freezing early layers~\cite{ye2023partial}, applying layer-wise learning rate decay~\cite{wortsman2022robust}, or updating only selected parameters such as Feed-Forward Network (FFN) or attention sub-layers.
They reduce representation drift but may sacrifice task performance if overly restrictive, and introduce a lot of extra hyperparameters.

Parameter-efficient fine-tuning (PEFT) methods such as Visual Prompt Tuning (VPT)~\cite{jia2022vpt}, adapters~\cite{chen2022adaptformer}, and Low-Rank Adaptation (LoRA)~\cite{hu2022lora} work towards the same goal by introducing minimal trainable parameters with the goal of trying to minimize the forgetting of the capabilities of the pretrained model. 

Most closely to our work are regularization methods, tightly related to the continual learning subfield, which constrain the deviation from the pretrained model's parameters~\cite{kirkpatrick2017overcoming} or its embeddings~\cite{cha2022domain}. 
LDIFS~\cite{mukhoti2024ldifs} and L2-SP~\cite{xuhong2018explicit} apply explicit losses to constrain fine-tuned features or weights to remain close to the pretrained model, respectively.
We adopt both of them in our work and study them when used in conjunction with fine-grained representation learning losses.
Different from previous work~\cite{mukhoti2024ldifs}, we show that for our task, regularization towards the pretrained parameters does not suffice in retaining the generic capabilities of the model.
In contrast to LDIFS~\cite{mukhoti2024ldifs}, we perform the embedding regularization on generic data, unrelated to the fine-tuning dataset's domain, and we also only use the last layer's embeddings for the regularization, as it would be prohibitive to use more in our larger-scale setting.
This resembles the work of~\cite{zheng2023preventing}, which, however, does not combine it with L2-SP and focuses on standard classification tasks in the continual learning setting. 
What additionally differentiates us from this work is that we proceed to evaluate two different and more modern VLMs, namely SigLIP and TIPS, instead of the CLIP model, and that we do not require the use of the text encoder or text data during training.
Anchor-based Robust Fine-Tuning (ARF)~\cite{han2024arf} introduced the idea of retrieving related examples from external generic datasets (e.g., LAION, CC3M) and enriching the fine-tuning process with semantically rich anchors. While ARF demonstrated improved out-of-distribution (OOD) robustness, it primarily focused on anchors semantically similar to the fine-tuning task. 
We show that such an idea is actually harmful in our setting, and the use of unrelated data to the fine-tuning domain at hand is the correct choice.

In another line of work, weight interpolation (averaging) methods such as WiSE-FT~\cite{wortsman2022robust} blend pretrained and fine-tuned model weights post hoc, balancing task specialization with retention of original capabilities. 
However, this requires maintaining multiple sets of weights and assumes that interpolation does not compromise either objective significantly.
We provide comparisons with WISE-FT and show that it underperforms our method on overall capabilities.

\vspace{-10pt}

\paragraph{Fine-grained representation learning.}
Fine-grained representation (metric) learning is most commonly cast as open-set retrieval, \ie, a model is trained on classes of a fine-grained domain (e.g., car models, mushroom species, etc.) and then tested on different classes of the same visual domain. 
Previous work focuses on maximizing the in-domain performance of models by developing better architectures~\cite{he2022transfg}, losses~\cite{ptm22,ramzi2025optimization}, or sample mining strategies~\cite{xuan2020improved,xuan2020hard}. 
They follow the standard path, initializing from strong pretrained models, as we do, but do not evaluate the impact of the fine-tuning process on the pretrained knowledge of the model. 

Some of the recent continual learning works, like that of~\cite{zheng2023preventing}, include fine-grained datasets in their evaluation.
However, they are evaluated in a different manner, namely, zero-shot classification that uses the textual description of the class.
In this work, we consider larger-scale and more realistic datasets, like iNaturalist, Food2k, etc., which span many orders of magnitude more samples and classes, and they are evaluated in an open set scenario.
For evaluating the retention of text-image capabilities, we instead use more realistic image-text evaluation datasets, like Flickr30k~\cite{young2014from} and COCO~\cite{chen2015microsoft}, instead of simply evaluating the zero-shot performance of the VLM, which includes fixed class names as the textual query.
\vspace{-10pt}

\paragraph{Generalization and selection of the validation set.}
The importance of a proper validation set that accurately predicts the final test performance of the model is highlighted in the recent work~\cite{etc25}, where the authors show that an improper validation set can fail to pick the correct hyperparameters of a given method, let alone the best method across competitors.
This is especially relevant to our work, where two contrasting objectives are being optimized, namely, in-domain performance attainment and generic knowledge retention. 
The choice of the mixing weights that balance the trade-off between the two objectives assumes a proper validation set that depicts this trade-off.
Unfortunately, previous work either uses in-domain data as a validation set~\cite{mukhoti2024ldifs}, which can not effectively, if at all, optimize both objectives at the same time, or does not reveal on which data (in or out-of-domain) the weights that trade-off in for out-of-domain performance are tuned on~\cite{zheng2023preventing}.
We strive to provide a proper way of choosing a validation set with the aforementioned properties, and a streamlined pipeline on how to choose the mixing weights.
\section{Fine-tuning the VLM} 
\label{sec:method}

In this section, we propose a fine-tuning pipeline designed explicitly to adapt the pretrained image embedding to the fine-grained visual domain while preserving as much knowledge as possible from the pretrained Vision-Language Model (VLM). 
Two distinct regularization approaches are thoroughly analyzed, and a combined loss function is developed that synergizes their complementary benefits. Furthermore, we emphasize the necessity of principled hyperparameter tuning and validation to achieve optimal performance with a justified methodology.

\subsection{Preliminaries}

\begin{figure}[tbp]
    \centering
    \hspace{-4pt}
    \scalebox{0.825}{
    \includegraphics[scale=0.215]{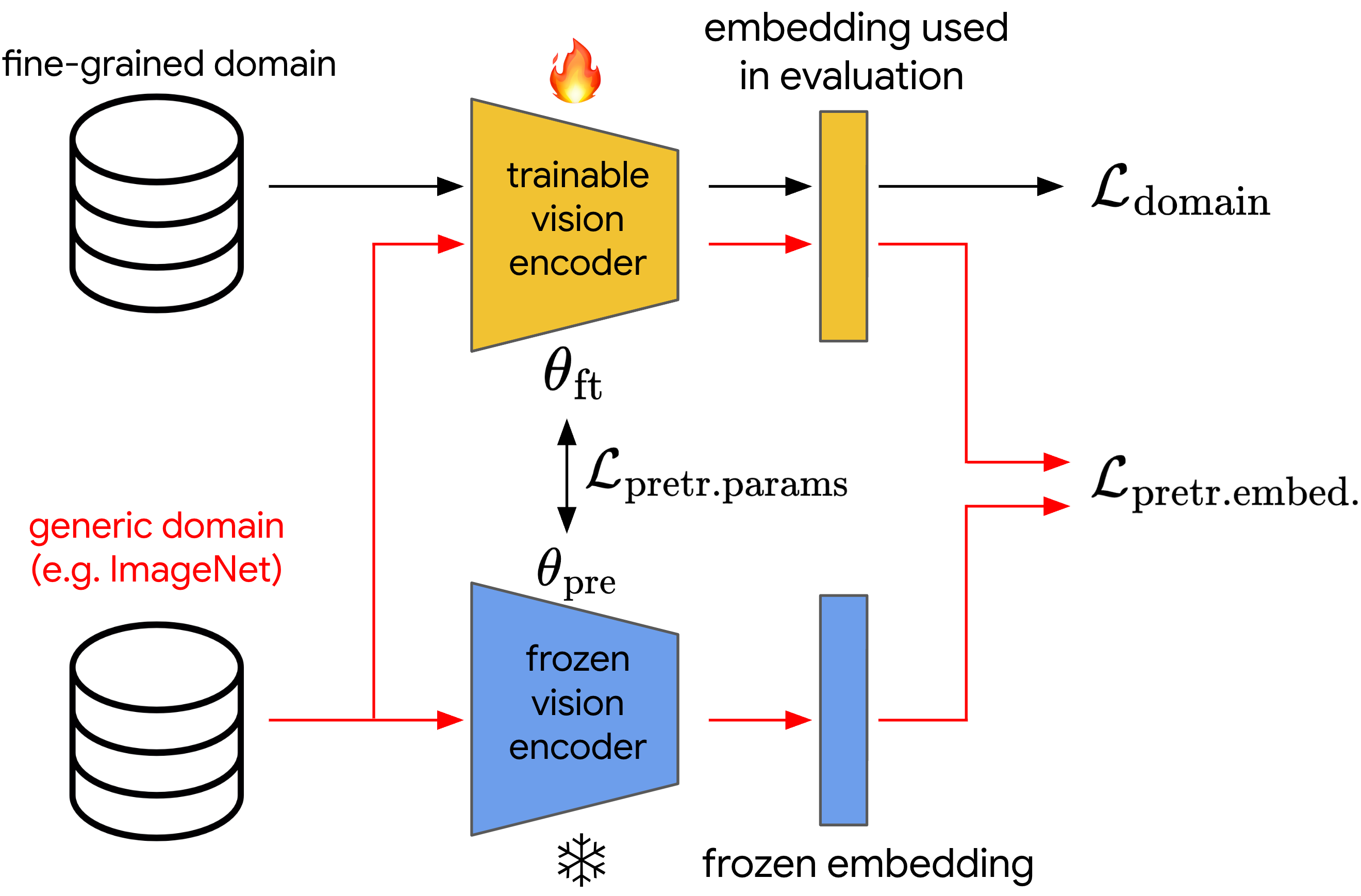}
    }
    \caption{Schematic illustration of the proposed fine-tuning pipeline. 
    The vision encoder that is being fine-tuned is initialized with the parameters of the frozen vision encoder of the VLM.
    During fine-tuning, it tries to minimize the in-domain loss on the fine-grained domain's data $(\mathcal{L}_{\text{domain}})$, while at the same time it is tasked to not drift far away from the pretrained parameters $(\mathcal{L}_{\text{pretr.params}})$ and the embedding space $(\mathcal{L}_{\text{pretr.embed.}})$ of the frozen vision encoder of the VLM.
    }
    \label{fig:method_example}

\vspace{-8pt}

\end{figure}

The image embedding is produced by the vision encoder of the Vision-Language Foundation model (typically implemented as a Vision Transformer (ViT)), which will be referred to as the ``backbone''.
In this work, the image embedding corresponds to the [CLS] token of the standard ViT architecture; however, other choices of the embedding and backbone architecture can also be utilized.
Let \( f_{\theta}: \mathcal{X} \rightarrow \mathbb{R}^D \) denote the \l2 normalized embedding output by the Vision Transformer as a function that takes an input image \( x \in \mathcal{X} \) and maps it to the \([\text{CLS}]\) token \( f_{\theta} \in \mathbb{R}^D \).
For the sake of notation compactness, we will use \( f(x) \) if the network parameter identification $\theta$ is clear from the context.
Notably, our fine-tuning pipeline exclusively leverages the visual encoder and entirely excludes the text encoder.
A schematic illustration of our approach is depicted in Figure~\ref{fig:method_example}.
The model to be fine-tuned (in orange color, top) is initialized with the parameters of the pretrained VLM's vision encoder, and produces the embedding that will acquire both fine-grained knowledge of the newly presented domain, but also retain the generic knowledge concentrated in the frozen encoder's embedding.

\subsection{Domain learning with class-level supervision}

To adapt the pretrained embedding to the fine-grained domain at hand, a classification-based representation learning approach is used.
More specifically, given the annotated fine-grained data (image-label pairs), fine-tuning of the embedding is achieved by appending a cosine classifier~\cite{zhai2018classification} on top of it, and backpropagating the commonly used normalized cross-entropy loss~\cite{zhai2018classification} through the embedding.
The loss function is defined as
\begin{equation}
\begin{split}
\Lcls 
  &= - \frac{1}{B} \sum_{j=1}^{B} y_{j} \log\!\left(\hat{y}_{j}\right) \\  
  &= - \frac{1}{B} \sum_{j=1}^{B} 
       \log\!\left(
         \frac{\exp\bigl(p_{y_j}^\top f_\theta(x_j)\bigr)}
              {\sum_{k=1}^{K}\exp\bigl(p_k^\top f_\theta(x_j)\bigr)}
       \right) \,.
\end{split}
\label{eqn:crossentropy}
\end{equation}

\noindent where $B$ is the batch size, \( y_{j} \) is the one-hot ground truth vector of sample $j$, \( \hat{y}_{j} \) is the softmax-ed predicted probability vector for sample $j$ produced by the classifier, $K$ is the number of classes and $p_{k}$ denotes the \l2 normalized learnable classifier prototype
of class $k$.
The number of classes $K$ is defined by the training set of the fine-grained domain at hand; note, however, that this classifier is discarded at test time, as the test classes are unknown, \ie the retrieval task is open set.

\pgfplotstableread{
epoch loss
1 0.532000
764 0.741000
1528 0.765000
2292 0.773000
3056 0.779000
3820 0.781000
4584 0.782000
5348 0.783000
6112 0.779000
6876 0.776000
7640 0.776000
8404 0.772000
9168 0.773000
9932 0.763000
10696 0.764000
11460 0.761000
12224 0.758000
12988 0.758000
13752 0.757000
14516 0.733000
15280 0.745000
16044 0.749000
16808 0.749000
17572 0.748000
18336 0.734000
19100 0.732000
19864 0.737000
20628 0.731000
21392 0.734000
22156 0.707000
22920 0.731000
23684 0.717000
24448 0.716000
25167 0.731000
}{\paramsregzeroweight}

\pgfplotstableread{
epoch loss
1 0.532000
764 0.741000
1528 0.765000
2292 0.776000
3056 0.776000
3820 0.780000
4584 0.780000
5348 0.784000
6112 0.780000
6876 0.780000
7640 0.781000
8404 0.776000
9168 0.776000
9932 0.779000
10696 0.775000
11460 0.772000
12224 0.765000
12988 0.763000
13752 0.767000
14516 0.767000
15280 0.764000
16044 0.752000
16808 0.762000
17572 0.763000
18336 0.761000
19100 0.755000
19864 0.755000
20628 0.757000
21392 0.750000
22156 0.756000
22920 0.752000
23684 0.747000
24448 0.749000
25167 0.752000
}{\paramsregtenthirdweight}

\pgfplotstableread{
epoch loss
1 0.532000
764 0.737000
1528 0.762000
2292 0.773000
3056 0.773000
3820 0.781000
4584 0.779000
5348 0.790000
6112 0.786000
6876 0.789000
7640 0.789000
8404 0.789000
9168 0.792000
9932 0.794000
10696 0.793000
11460 0.794000
12224 0.790000
12988 0.795000
13752 0.793000
14516 0.787000
15280 0.792000
16044 0.791000
16808 0.791000
17572 0.792000
18336 0.788000
19100 0.793000
19864 0.788000
20628 0.787000
21392 0.792000
22156 0.788000
22920 0.791000
23684 0.790000
24448 0.791000
25167 0.789000
}{\paramsregtenfourthweight}

\pgfplotstableread{
epoch loss
1 0.532000
764 0.723000
1528 0.751000
2292 0.759000
3056 0.772000
3820 0.773000
4584 0.780000
5348 0.784000
6112 0.786000
6876 0.788000
7640 0.786000
8404 0.788000
9168 0.791000
9932 0.792000
10696 0.792000
11460 0.793000
12224 0.795000
12988 0.789000
13752 0.792000
14516 0.794000
15280 0.795000
16044 0.794000
16808 0.797000
17572 0.795000
18336 0.795000
19100 0.795000
19864 0.796000
20628 0.797000
21392 0.794000
22156 0.795000
22920 0.797000
23684 0.798000
24448 0.799000
25167 0.796000
}{\paramsregtenfifthweight}

\pgfplotstableread{
epoch loss
1 0.532000
764 0.687000
1528 0.710000
2292 0.717000
3056 0.727000
3820 0.731000
4584 0.737000
5348 0.741000
6112 0.745000
6876 0.746000
7640 0.750000
8404 0.746000
9168 0.752000
9932 0.753000
10696 0.753000
11460 0.752000
12224 0.752000
12988 0.752000
13752 0.751000
14516 0.752000
15280 0.751000
16044 0.756000
16808 0.755000
17572 0.753000
18336 0.754000
19100 0.753000
19864 0.748000
20628 0.754000
21392 0.754000
22156 0.753000
22920 0.757000
23684 0.756000
24448 0.755000
25167 0.755000
}{\paramsregtensixthweight}

\pgfplotstableread{
epoch loss
1 0.826000
764 0.804000
1528 0.791000
2292 0.785000
3056 0.777000
3820 0.770000
4584 0.771000
5348 0.760000
6112 0.751000
6876 0.747000
7640 0.740000
8404 0.740000
9168 0.731000
9932 0.726000
10696 0.726000
11460 0.724000
12224 0.721000
12988 0.715000
13752 0.707000
14516 0.702000
15280 0.695000
16044 0.695000
16808 0.692000
17572 0.692000
18336 0.693000
19100 0.689000
19864 0.679000
20628 0.679000
21392 0.685000
22156 0.681000
22920 0.671000
23684 0.677000
24448 0.669000
25167 0.662000
}{\paramsregzeroweightood}

\pgfplotstableread{
epoch loss
1 0.826000
764 0.806000
1528 0.795000
2292 0.788000
3056 0.780000
3820 0.776000
4584 0.775000
5348 0.766000
6112 0.764000
6876 0.759000
7640 0.764000
8404 0.742000
9168 0.742000
9932 0.746000
10696 0.741000
11460 0.745000
12224 0.739000
12988 0.726000
13752 0.731000
14516 0.740000
15280 0.734000
16044 0.710000
16808 0.716000
17572 0.730000
18336 0.720000
19100 0.721000
19864 0.699000
20628 0.718000
21392 0.699000
22156 0.709000
22920 0.693000
23684 0.692000
24448 0.686000
25167 0.703000
}{\paramsregtenthirdweightood}

\pgfplotstableread{
epoch loss
1 0.826000
764 0.808000
1528 0.795000
2292 0.789000
3056 0.784000
3820 0.781000
4584 0.777000
5348 0.769000
6112 0.767000
6876 0.759000
7640 0.760000
8404 0.746000
9168 0.751000
9932 0.743000
10696 0.739000
11460 0.745000
12224 0.754000
12988 0.737000
13752 0.731000
14516 0.746000
15280 0.738000
16044 0.732000
16808 0.733000
17572 0.734000
18336 0.734000
19100 0.733000
19864 0.732000
20628 0.731000
21392 0.731000
22156 0.727000
22920 0.714000
23684 0.729000
24448 0.721000
25167 0.737000
}{\paramsregtenfourthweightood}

\pgfplotstableread{
epoch loss
1 0.826000
764 0.808000
1528 0.804000
2292 0.802000
3056 0.796000
3820 0.790000
4584 0.792000
5348 0.785000
6112 0.783000
6876 0.784000
7640 0.783000
8404 0.788000
9168 0.781000
9932 0.785000
10696 0.785000
11460 0.781000
12224 0.775000
12988 0.778000
13752 0.780000
14516 0.786000
15280 0.777000
16044 0.775000
16808 0.780000
17572 0.782000
18336 0.779000
19100 0.783000
19864 0.789000
20628 0.788000
21392 0.786000
22156 0.790000
22920 0.778000
23684 0.779000
24448 0.785000
25167 0.783000
}{\paramsregtenfifthweightood}

\pgfplotstableread{
epoch loss
1 0.826000
764 0.817000
1528 0.816000
2292 0.813000
3056 0.809000
3820 0.806000
4584 0.808000
5348 0.806000
6112 0.800000
6876 0.804000
7640 0.801000
8404 0.804000
9168 0.803000
9932 0.802000
10696 0.805000
11460 0.802000
12224 0.802000
12988 0.803000
13752 0.803000
14516 0.803000
15280 0.801000
16044 0.805000
16808 0.800000
17572 0.801000
18336 0.802000
19100 0.803000
19864 0.803000
20628 0.805000
21392 0.802000
22156 0.808000
22920 0.802000
23684 0.797000
24448 0.803000
25167 0.801000
}{\paramsregtensixthweightood}

\pgfplotstableread{
epoch loss
1 0.532000
764 0.741000
1528 0.765000
2292 0.773000
3056 0.779000
3820 0.781000
4584 0.782000
5348 0.783000
6112 0.779000
6876 0.776000
7640 0.776000
8404 0.772000
9168 0.773000
9932 0.763000
10696 0.764000
11460 0.761000
12224 0.758000
12988 0.758000
13752 0.757000
14516 0.733000
15280 0.745000
16044 0.749000
16808 0.749000
17572 0.748000
18336 0.734000
19100 0.732000
19864 0.737000
20628 0.731000
21392 0.734000
22156 0.707000
22920 0.731000
23684 0.717000
24448 0.716000
25167 0.731000
}{\embeddregzeroweight}

\pgfplotstableread{
epoch loss
1 0.532000
764 0.738000
1528 0.762000
2292 0.770000
3056 0.770000
3820 0.770000
4584 0.773000
5348 0.776000
6112 0.767000
6876 0.761000
7640 0.757000
8404 0.761000
9168 0.761000
9932 0.755000
10696 0.748000
11460 0.741000
12224 0.745000
12988 0.737000
13752 0.732000
14516 0.725000
15280 0.727000
16044 0.726000
16808 0.717000
17572 0.717000
18336 0.712000
19100 0.692000
19864 0.705000
20628 0.709000
21392 0.709000
22156 0.714000
22920 0.705000
23684 0.698000
24448 0.684000
25167 0.700000
}{\embeddregtensecondweight}

\pgfplotstableread{
epoch loss
1 0.532000
764 0.730000
1528 0.752000
2292 0.761000
3056 0.755000
3820 0.756000
4584 0.755000
5348 0.760000
6112 0.749000
6876 0.752000
7640 0.748000
8404 0.731000
9168 0.734000
9932 0.742000
10696 0.732000
11460 0.730000
12224 0.754000
12988 0.722000
13752 0.729000
14516 0.725000
15280 0.719000
16044 0.716000
16808 0.712000
17572 0.717000
18336 0.713000
19100 0.714000
19864 0.693000
20628 0.701000
21392 0.714000
22156 0.712000
22920 0.714000
23684 0.702000
24448 0.695000
25167 0.687000
}{\embeddregtenthirdweight}

\pgfplotstableread{
epoch loss
1 0.532000
764 0.701000
1528 0.725000
2292 0.726000
3056 0.734000
3820 0.732000
4584 0.729000
5348 0.730000
6112 0.737000
6876 0.726000
7640 0.735000
8404 0.731000
9168 0.732000
9932 0.731000
10696 0.730000
11460 0.741000
12224 0.741000
12988 0.721000
13752 0.729000
14516 0.733000
15280 0.732000
16044 0.746000
16808 0.728000
17572 0.704000
18336 0.723000
19100 0.716000
19864 0.717000
20628 0.721000
21392 0.722000
22156 0.718000
22920 0.689000
23684 0.709000
24448 0.731000
25167 0.721000
}{\embeddregtenfourthweight}

\pgfplotstableread{
epoch loss
1 0.532000
764 0.615000
1528 0.642000
2292 0.657000
3056 0.661000
3820 0.670000
4584 0.682000
5348 0.687000
6112 0.687000
6876 0.689000
7640 0.687000
8404 0.700000
9168 0.704000
9932 0.700000
10696 0.698000
11460 0.713000
12224 0.715000
12988 0.707000
13752 0.710000
14516 0.718000
15280 0.717000
16044 0.706000
16808 0.717000
17572 0.723000
18336 0.720000
19100 0.713000
19864 0.704000
20628 0.682000
21392 0.709000
22156 0.708000
22920 0.720000
23684 0.701000
24448 0.725000
25167 0.712000
}{\embeddregtenfifthweight}

\pgfplotstableread{
epoch loss
1 0.826000
764 0.804000
1528 0.791000
2292 0.785000
3056 0.777000
3820 0.770000
4584 0.771000
5348 0.760000
6112 0.751000
6876 0.747000
7640 0.740000
8404 0.740000
9168 0.731000
9932 0.726000
10696 0.726000
11460 0.724000
12224 0.721000
12988 0.715000
13752 0.707000
14516 0.702000
15280 0.695000
16044 0.695000
16808 0.692000
17572 0.692000
18336 0.693000
19100 0.689000
19864 0.679000
20628 0.679000
21392 0.685000
22156 0.681000
22920 0.671000
23684 0.677000
24448 0.669000
25167 0.662000
}{\embeddregzeroweightood}

\pgfplotstableread{
epoch loss
1 0.826000
764 0.807000
1528 0.799000
2292 0.795000
3056 0.796000
3820 0.798000
4584 0.795000
5348 0.787000
6112 0.794000
6876 0.798000
7640 0.798000
8404 0.799000
9168 0.802000
9932 0.801000
10696 0.794000
11460 0.795000
12224 0.799000
12988 0.795000
13752 0.795000
14516 0.788000
15280 0.799000
16044 0.797000
16808 0.795000
17572 0.797000
18336 0.797000
19100 0.795000
19864 0.800000
20628 0.797000
21392 0.790000
22156 0.796000
22920 0.800000
23684 0.800000
24448 0.794000
25167 0.796000
}{\embeddregtensecondweightood}

\pgfplotstableread{
epoch loss
1 0.826000
764 0.814000
1528 0.811000
2292 0.811000
3056 0.814000
3820 0.813000
4584 0.815000
5348 0.813000
6112 0.816000
6876 0.817000
7640 0.816000
8404 0.819000
9168 0.816000
9932 0.819000
10696 0.817000
11460 0.821000
12224 0.809000
12988 0.817000
13752 0.818000
14516 0.814000
15280 0.817000
16044 0.815000
16808 0.816000
17572 0.817000
18336 0.821000
19100 0.820000
19864 0.814000
20628 0.815000
21392 0.816000
22156 0.816000
22920 0.818000
23684 0.817000
24448 0.817000
25167 0.815000
}{\embeddregtenthirdweightood}

\pgfplotstableread{
epoch loss
1 0.826000
764 0.821000
1528 0.821000
2292 0.823000
3056 0.820000
3820 0.821000
4584 0.822000
5348 0.822000
6112 0.817000
6876 0.822000
7640 0.818000
8404 0.821000
9168 0.820000
9932 0.820000
10696 0.821000
11460 0.822000
12224 0.821000
12988 0.820000
13752 0.818000
14516 0.821000
15280 0.818000
16044 0.820000
16808 0.820000
17572 0.817000
18336 0.817000
19100 0.817000
19864 0.821000
20628 0.820000
21392 0.820000
22156 0.820000
22920 0.818000
23684 0.820000
24448 0.816000
25167 0.817000
}{\embeddregtenfourthweightood}

\pgfplotstableread{
epoch loss
1 0.826000
764 0.822000
1528 0.823000
2292 0.825000
3056 0.820000
3820 0.820000
4584 0.818000
5348 0.823000
6112 0.820000
6876 0.824000
7640 0.822000
8404 0.820000
9168 0.822000
9932 0.820000
10696 0.821000
11460 0.822000
12224 0.822000
12988 0.820000
13752 0.818000
14516 0.821000
15280 0.819000
16044 0.820000
16808 0.822000
17572 0.818000
18336 0.819000
19100 0.818000
19864 0.820000
20628 0.822000
21392 0.819000
22156 0.818000
22920 0.821000
23684 0.820000
24448 0.818000
25167 0.820000
}{\embeddregtenfifthweightood}

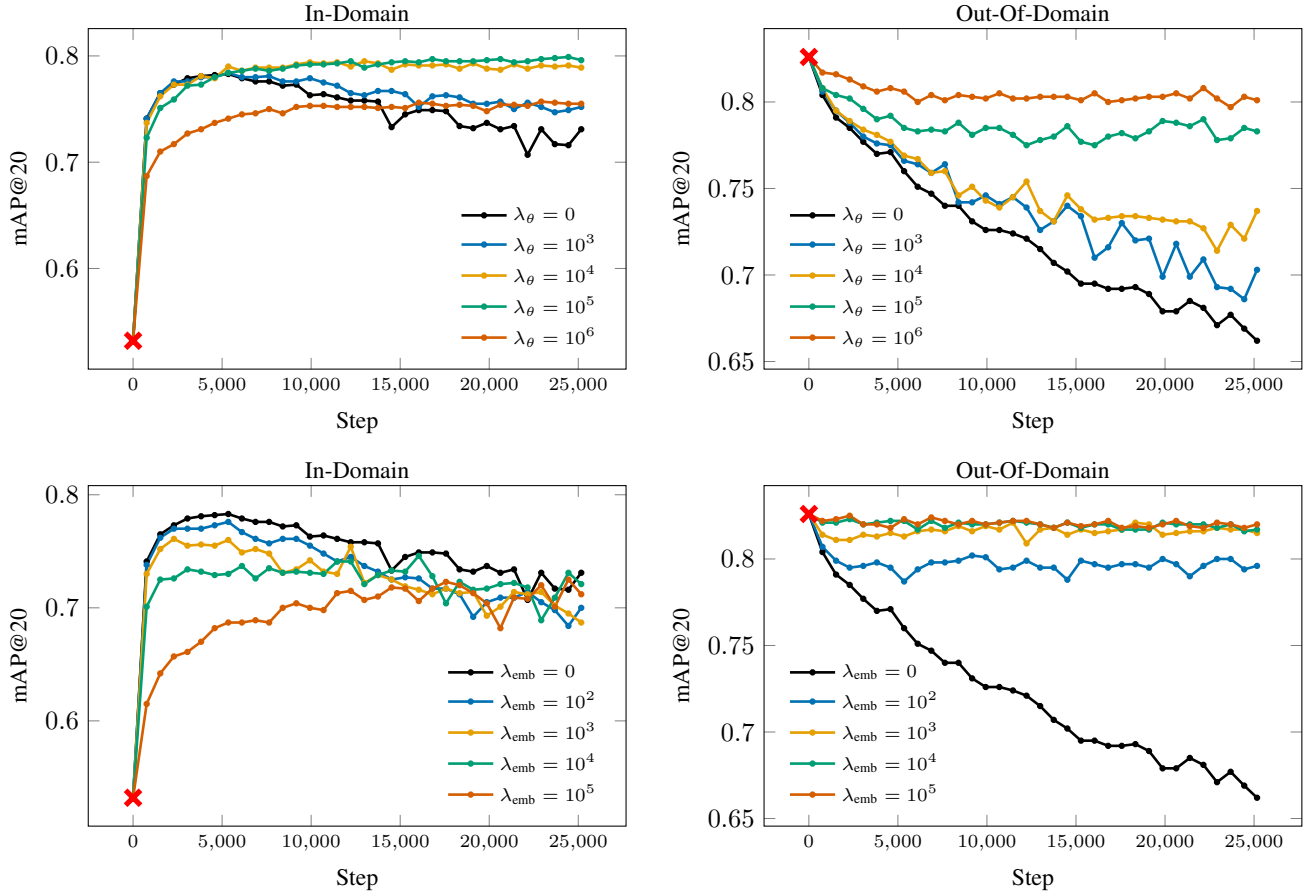
\begin{figure*}[ht]
    \centering
\begin{tikzpicture}

\begin{axis}[
    name=inplot,
    title={In-Domain},
    title style={
        font=\small,
        yshift=-1.5ex
    },
    width=0.5\linewidth,
    height=0.35\linewidth,
    xlabel={\small Step},
    ylabel={\small mAP@20},
    scaled x ticks=false,
    xticklabel style={font=\scriptsize},  
    legend pos=south east,
    legend style={
        font=\scriptsize,
        draw=none,
        fill=none,
        align=left,
        cells={anchor=west},
    },
]

    \addplot[color=black,solid, mark=*,  mark size=0.7, line width=1.0] table[x=epoch, y expr={\thisrow{loss}}]\paramsregzeroweight 
    ;
    \addlegendentry{$\lamtheta = 0$}

    \addplot[color=tolblue,solid, mark=*,  mark size=0.7, line width=1.0] table[x=epoch, y expr={\thisrow{loss}}]\paramsregtenthirdweight 
    ;
    \addlegendentry{$\lamtheta = 10^3$}

    \addplot[color=tolorange,solid, mark=*,  mark size=0.7, line width=1.0] table[x=epoch, y expr={\thisrow{loss}}]\paramsregtenfourthweight 
    ;
    \addlegendentry{$\lamtheta = 10^4$}
    
    \addplot[color=tolgreen,solid, mark=*,  mark size=0.7, line width=1.0] table[x=epoch, y expr={\thisrow{loss}}]\paramsregtenfifthweight 
    ;
    \addlegendentry{$\lamtheta = 10^5$}
    
    \addplot[color=tolred,solid, mark=*,  mark size=0.7, line width=1.0] table[x=epoch, y expr={\thisrow{loss}}]\paramsregtensixthweight 
    ;
    \addlegendentry{$\lamtheta = 10^6$}

    \addplot[
        color=red,
        mark=x,
        mark size=4,
        only marks,
        line width=2pt
    ] coordinates {(1, 0.532)};

\end{axis}

\begin{axis}[
    name=outplot,
    title={Out-Of-Domain},
    title style={
        font=\small,
        yshift=-1.5ex 
    },
    at={(inplot.right of south east)},
    anchor=left of south west,
    xshift=0.5cm,
    width=0.5\linewidth,
    height=0.35\linewidth,
    xlabel={\small Step},
    ylabel={\small mAP@20},
    scaled x ticks=false,
    xticklabel style={font=\scriptsize},
    legend pos=south west,
    legend style={
        font=\scriptsize,
        draw=none,
        fill=none,
        align=left,
        cells={anchor=west},
    },
]

    \addplot[color=black,solid, mark=*,  mark size=0.7, line width=1.0] table[x=epoch, y expr={\thisrow{loss}}]\paramsregzeroweightood 
    ;
    \addlegendentry{$\lamtheta = 0$}
    
    \addplot[color=tolblue,solid, mark=*,  mark size=0.7, line width=1.0] table[x=epoch, y expr={\thisrow{loss}}]\paramsregtenthirdweightood 
    ;
    \addlegendentry{$\lamtheta = 10^3$}
        
    \addplot[color=tolorange,solid, mark=*,  mark size=0.7, line width=1.0] table[x=epoch, y expr={\thisrow{loss}}]\paramsregtenfourthweightood 
    ;
    \addlegendentry{$\lamtheta = 10^4$}
    
    \addplot[color=tolgreen,solid, mark=*,  mark size=0.7, line width=1.0] table[x=epoch, y expr={\thisrow{loss}}]\paramsregtenfifthweightood 
    ;
    \addlegendentry{$\lamtheta = 10^5$}
    
    \addplot[color=tolred,solid, mark=*,  mark size=0.7, line width=1.0] table[x=epoch, y expr={\thisrow{loss}}]\paramsregtensixthweightood 
    ;
    \addlegendentry{$\lamtheta = 10^6$}

    \addplot[
        color=red,
        mark=x,
        mark size=4,
        only marks,
        line width=2pt
    ] coordinates {(1, 0.826)};

\end{axis}

\begin{axis}[
    name=inplot2,
    title={In-Domain},
    title style={
        font=\small,
        yshift=-1.5ex 
    },
    at={(inplot.south west)},
    anchor=north west,
    yshift=-1.55cm, 
    width=0.5\linewidth,
    height=0.35\linewidth,
    xlabel={\small Step},
    ylabel={\small mAP@20},
    scaled x ticks=false,
    xticklabel style={font=\scriptsize},
    legend pos=south east,
    legend style={
        font=\scriptsize,
        draw=none,
        fill=none,
        align=left,
        cells={anchor=west},
    },
]

    \addplot[color=black,solid, mark=*,  mark size=0.7, line width=1.0] table[x=epoch, y expr={\thisrow{loss}}]\embeddregzeroweight 
    ;
    \addlegendentry{$\lameb = 0$}
    
    \addplot[color=tolblue,solid, mark=*,  mark size=0.7, line width=1.0] table[x=epoch, y expr={\thisrow{loss}}]\embeddregtensecondweight 
    ;
    \addlegendentry{$\lameb = 10^2$}
    
    \addplot[color=tolorange,solid, mark=*,  mark size=0.7, line width=1.0] table[x=epoch, y expr={\thisrow{loss}}]\embeddregtenthirdweight 
    ;
    \addlegendentry{$\lameb = 10^3$}
    
    \addplot[color=tolgreen,solid, mark=*,  mark size=0.7, line width=1.0] table[x=epoch, y expr={\thisrow{loss}}]\embeddregtenfourthweight 
    ;
    \addlegendentry{$\lameb = 10^4$}
    
    \addplot[color=tolred,solid, mark=*,  mark size=0.7, line width=1.0] table[x=epoch, y expr={\thisrow{loss}}]\embeddregtenfifthweight 
    ;
    \addlegendentry{$\lameb = 10^5$}

    \addplot[
        color=red,
        mark=x,
        mark size=4,
        only marks,
        line width=2pt
    ] coordinates {(1, 0.532)};

\end{axis}

\begin{axis}[%
    name=outplot2,
    title={Out-Of-Domain},
    title style={
        font=\small,
        yshift=-1.5ex  
    },
    at={(outplot.south west)},
    anchor=north west,
    yshift=-1.55cm, 
    width=0.5\linewidth,
    height=0.35\linewidth,
    xlabel={\small Step},
    ylabel={\small mAP@20},
    scaled x ticks=false,
    xticklabel style={font=\scriptsize},     legend pos=south west,
    legend style={
        font=\scriptsize,
        draw=none,
        fill=none,
        align=left,
        cells={anchor=west},
    },
]

    \addplot[color=black,solid, mark=*,  mark size=0.7, line width=1.0] table[x=epoch, y expr={\thisrow{loss}}]\embeddregzeroweightood 
    ;
    \addlegendentry{$\lameb = 0$}
    
    \addplot[color=tolblue,solid, mark=*,  mark size=0.7, line width=1.0] table[x=epoch, y expr={\thisrow{loss}}]\embeddregtensecondweightood 
    ;
    \addlegendentry{$\lameb = 10^2$}
    
    \addplot[color=tolorange,solid, mark=*,  mark size=0.7, line width=1.0] table[x=epoch, y expr={\thisrow{loss}}]\embeddregtenthirdweightood 
    ;
    \addlegendentry{$\lameb = 10^3$}
    
    \addplot[color=tolgreen,solid, mark=*,  mark size=0.7, line width=1.0] table[x=epoch, y expr={\thisrow{loss}}]\embeddregtenfourthweightood 
    ;
    \addlegendentry{$\lameb = 10^4$}
    
    \addplot[color=tolred,solid, mark=*,  mark size=0.7, line width=1.0] table[x=epoch, y expr={\thisrow{loss}}]\embeddregtenfifthweightood 
    ;
    \addlegendentry{$\lameb = 10^5$}

    \addplot[
        color=red,
        mark=x,
        mark size=4,
        only marks,
        line width=2pt
    ] coordinates {(1, 0.826)};

\end{axis}

\end{tikzpicture}

\caption{
Validation performance (mAP@20) on in-domain (SOP) and out-of-domain (Food2k) data during fine-tuning of SigLIP ViT-Base/16 on the SOP training set. 
The top row shows results with different values of the parameter regularization weight $\lamtheta$, while the bottom row shows results for the embedding regularization weight $\lameb$.
Only one regularization is applied at a time; the other is set to zero. 
Colored lines represent different regularization strengths (as indicated in the legend), and the red cross marks the performance of the pretrained model before fine-tuning.
}
\label{fig:study}
\vspace{-8pt}

\end{figure*}

\subsection{Regularization towards pretrained knowledge}
The classification loss ensures that the model learns the specifics of the fine-grained domain that it is being fine-tuned on. 
By altering the model embedding to discriminate the training classes, the knowledge encoded in the pretrained VLM is being erased.
In order to mitigate that effect, we introduce a combination of two types of regularization to try to preserve the embedding structure with the semantics of the multimodal space of the pretrained model.

\paragraph{Parameter regularization.} 
This loss is independent of the input examples and pushes the internal parameters $\theta_{\text{ft}}$ of the fine-tuned network towards the parameters of the pre-trained network $\theta_{\text{pre}}$.
Proposed in~\cite{xuhong2018explicit}, the L2-SP regularization is defined as
\begin{equation}
     \Ltheta = \frac{1}{N} \sum_{i=1}^{N} \left\| \theta_{\text{ft}}^{(i)} - \theta_{\text{pre}}^{(i)} \right\|_2^2 \mbox{,}
\end{equation}
where the $N$ indexes the network parameters.

\paragraph{Embedding regularization.} 
To explicitly constrain embedding drift, an LDIFS~\cite{mukhoti2024ldifs}-inspired regularization is employed, which distills the backbone embeddings of the pretrained network (target embeddings) to the backbone embeddings of the network that is being fine-tuned:
\begin{equation}
    \Leb = \frac{1}{B} \sum_{j=1}^{B} \left\| f_{\theta_{\text{ft}}}(x_j) - f_{\theta_{\text{pre}}}(x_j) \right\|_2^2 \mbox{,}
\end{equation}
where $B$ is the batch size, and $j$ indexes samples of the batch.
For our method, this loss is only applied to images from an external generic dataset that is not related to the domain being trained. 
In particular, we choose to use ImageNet~\cite{russakovsky2015imagenet}
due to its ease of access and widespread use. 
The batches of ImageNet are fed in a round-robin fashion, interleaved between batches of the fine-tuning domain at hand.
In practice, no forward pass through the pretrained (frozen) vision encoder that produces the distillation targets is done during training; instead, the embeddings are extracted offline once before training, saving compute and time.

\paragraph{The objective function.}
The final loss function is a weighted combination of the three losses
\begin{equation} \label{eqn:total}
    \mathcal{L}_{\text{total}} = \mathcal{L}_{\text{domain}} +  \lameb \mathcal{L}_{\text{pretr.embed.}} +  \lamtheta \mathcal{L}_{\text{pretr.params}} ~\mbox{.}
\end{equation}
In the following sections, we deliver intuitive interpretation of the two regularizaton losses and propose a method of setting the weights $\lameb$  and $\lamtheta$.

\subsection{Regularization functionality} 

To motivate our proposed fine-tuning method, we begin by evaluating each regularization term independently to illustrate its respective effects on both in-domain adaptation and out-of-domain generalization. 
We use a running example of fine-tuning on the SOP~\cite{song2016deep} dataset (domain of online product images) and evaluation on Food2k~\cite{min2023large} (domain of food images) as out-of-domain, see Fig~\ref{fig:study}. Note that the values shown in the out-of-domain data are \emph{never} available or used during training - it is only used here as an illustrative example, to justify the fine-tuning technique we introduce.

\paragraph{Parameter regularization.} In this paragraph, the impact of the weight $\lamtheta$ in the loss function (eqn.~\ref{eqn:total}), while $\lameb = 0$, is studied.
Figure~\ref{fig:study} (top row panels) illustrates the impact of different weights for the parameter regularization loss. 
Fine-tuning with minimal or no regularization rapidly improves in-domain performance, but eventually leads to overfitting,
as indicated by the subsequent decline on the SOP validation set. Introducing moderate parameter regularization (e.g., weight $10^4$) effectively curbs this overfitting, resulting in more stable in-domain accuracy throughout training. 
However, as the regularization strength increases further, adaptation to the fine-grained domain becomes increasingly constrained.
From the out-of-domain perspective (Figure~\ref{fig:study}, top row panels, right panel),
parameter regularization alone does not suffice to retain performance on unrelated domains: regardless of the regularization strength (unless it is so large that the network does not adapt at all), the embedding distribution can still drift away from the pretrained one. This leads to a rapid drop in out-of-domain accuracy, highlighting a limitation of constraining only the network parameters.
This comes in contrast to conclusions from previous work~\cite{mukhoti2024ldifs}, where the parameter regularization seemed to be effective. We hypothesize this has to do with the different nature of the fine-tuning tasks.

\paragraph{Embedding regularization.}
Embedding regularization, in contrast, explicitly restricts the drift of the learned representation from the pretrained embedding, but comes with its own trade-offs. As shown in Figure~\ref{fig:study} (bottom row panels, right panel), sufficiently strong embedding regularization can nearly freeze out-of-domain performance at its initial value, effectively preventing catastrophic forgetting. 
However, this comes at the cost of reduced adaptation to the fine-grained domain: increasing the embedding loss weight consistently degrades in-domain performance (bottom row panels, left panel).

\subsection{Loss weights from joint validation} 
\label{valid}
The final loss function (eqn.~\ref{eqn:total}) is a combination of the three losses.
The effectiveness of our fine-tuning pipeline ultimately depends on finding an appropriate balance between domain adaptation and knowledge preservation. This balance is controlled by the weights $(\lamtheta,\lameb)$ assigned to the parameter and embedding regularization terms, respectively. 
In practice, these hyperparameters can have a substantial impact on both in-domain and out-of-domain performance and require careful tuning.
If only in-domain validation is used for model selection or early stopping, the resulting model may exhibit strong adaptation but poor retention of pretrained knowledge, or vice versa. 
To address this, we adopt a validation strategy with dual components: 
\begin{itemize}
    \item For the in-domain component, the standard validation set of the fine-grained dataset is used, measuring the relevant retrieval metric.
    \item For the out-of-domain component, a subset of a large, generic dataset is reserved, in our case, a held-out portion of the ImageNet training set, and used with the same metric to monitor performance.
\end{itemize}
During training, the final validation score is computed as the average of the in-domain and out-of-domain metrics. 
This composite metric is used both for hyperparameter tuning (to select the optimal weights for the regularization losses) and for early stopping. 
By optimizing for the average rather than the fine-tuning domain only, we explicitly encourage the model to strike a balance between adapting to the new domain and retaining its generic pretrained knowledge.
This validation protocol is crucial for achieving reliable and reproducible results. 
Without an out-of-domain validation set, one selects models that are overfit to the fine-tuning domain and have lost most of their generalization ability, a limitation that is often overlooked in prior work. 
\section{Experiments}
\label{sec:experiments}

\begin{table*}[htbp]
    \centering
    \scalebox{0.98}{
    \begin{tabular}{c|c|c||c|c|c|c}
    \hline
    
            Fine-tuning & Fine-tuning & Reg. weights & \multicolumn{2}{c|}{Image-Image (mAP@20)} & \multicolumn{1}{c|}{Text-Img (R@1)} \\ 
     Domain & Method  & $(\lameb,\lamtheta)$   & In-Domain & O-o-D Average & O-o-D Average & In-Out Avg. \\ \hline
                          
    \multicolumn{6}{c}{\textbf{SigLIP Pretraining}} \\ \hline
    \multirow{2}{*}{CARS} & Standard & $(-,-)$ & 91.5 (+5.5) & 43.2 (+0.3) & 65.4 (-3.0) & 72.9 (+2.1) \\
                          & Ours & $(10^2,10^4)$ & \textbf{92.6 (+6.6)} & \textbf{43.7 (+0.8)} & \textbf{67.4 (-1.0)} & \textbf{74.1 (+3.3)}\\
    \cline{1-7}
    \multirow{2}{*}{InShop} & Standard & $(-,-)$ & \textbf{85.4 (+33.5)} & \textbf{49.6 (-0.1)}   & 61.5 (-6.9) & 70.5 (+15.0)\\
                            & Ours & $(10^3,10^3)$ & 85.3 (+33.4) & 49.5 (-0.2) & \textbf{67.9 (-0.6)} & \textbf{72.0 (+16.5)} \\
    \cline{1-7}
    \multirow{2}{*}{SOP} & Standard & $(-,-)$& \textbf{76.9 (+31.0)} &  44.9 (-6.0) & 55.7 (-12.8) & 63.6 (+10.8) \\
                         & Ours & $(10^3,10^4)$ & 75.8 (+29.9) & \textbf{50.8 (-0.1)} & \textbf{67.5 (-0.9)} & \textbf{67.5 (+14.7)}\\          
    \cline{1-7}
    \multirow{2}{*}{iNat} & Standard & $(-,-)$& 57.4 (+14.8) & 49.9 (-1.6)  & 51.9 (-16.4) & 54.2 (+2.9)\\
                          & Ours & $(10^4,10^3)$& \textbf{57.5 (+14.9)} & \textbf{51.1 (-0.5)} & \textbf{66.1 (-2.4)} & \textbf{58.0 (+6.7)} \\
    \cline{1-7}
    \multirow{2}{*}{Food2k} & Standard & $(-,-)$& 48.8 (+20.4) &  46.7 (-7.7) & 55.7 (-12.8) & 50.0 (+5.1)\\
                            & Ours & $(10^4,10^3)$& \textbf{50.7 (+22.3)} & \textbf{53.8 (-0.6)} & \textbf{67.7 (-0.7)} & \textbf{55.7 (+10.8)}\\
    
    \hline
    \multicolumn{6}{c}{\textbf{TIPS Pretraining}} \\ \hline
    \multirow{2}{*}{CARS} & Standard& $(-,-)$ & 87.2 (+19.7) & 45.1 (+0.2) & 59.8 (-4.3) & 69.8 (+8.8)\\
                          & Ours& $(10^2,10^3)$ & \textbf{87.4 (+19.9)} & \textbf{45.4 (+0.6)} & \textbf{63.0 (-1.1)} & \textbf{70.8 (+9.8)} \\
    \cline{1-7}
    \multirow{2}{*}{InShop} & Standard & $(-,-)$& \textbf{85.1 (+30.6)} & \textbf{47.2 (-0.3)} & 53.2 (-10.9) & 67.6 (+12.5)\\
                            & Ours & $(10^3,10^3)$& 84.1 (+29.6) & \textbf{47.2 (-0.3)} & \textbf{63.6 (-0.5)} & \textbf{69.8 (+14.6)}\\
    \cline{1-7}
    \multirow{2}{*}{SOP} & Standard & $(-,-)$& \textbf{79.1 (+31.8)} & 41.2 (-7.7) & 45.8 (-18.3) & 61.3 (+9.4)\\
                         & Ours & $(10^2,10^3)$& 76.6 (+29.3) & \textbf{47.6 (-1.3)} & \textbf{63.7 (-0.4)} & \textbf{66.1 (+14.2)} \\
                         \cline{1-7}
    \multirow{2}{*}{iNat} & Standard & $(-,-)$& \textbf{65.6} (+19.2) & 44.3 (-4.8) & 28.6 (-35.6) & 51.0 (-0.5) \\
                          & Ours & $(10^2,10^3)$& 63.6 (+17.2) & \textbf{49.6 (+0.4)} & \textbf{59.3 (-4.8)} & \textbf{59.0 (+7.5)} \\
    \cline{1-7}
    \multirow{2}{*}{Food2k} & Standard & $(-,-)$& \textbf{52.3 (+20.4)} & 50.6 (-1.3) & 48.7 (-15.4) & 51.0 (+6.0)\\
                            & Ours & $(10^2,10^4)$& 50.7 (+18.8) & \textbf{52.3 (+0.3)} & \textbf{62.6 (-1.5)} & \textbf{54.1 (+9.1)}\\
    
    \hline
    
    \end{tabular}
    }
    \caption{
    Performance comparison of our method against standard fine-tuning, for different combinations of fine-tuning datasets and pre-trainings.
    The backbone used for all experiments is ViT-Base/16. 
    The performance difference ($\Delta$) to the pretrained model is reported next to the absolute performance.
    }
    \label{tab:main_table_all}

\vspace{-8pt}
    
\end{table*}

In this section, we conduct a thorough experimental evaluation to assess the benefits of the proposed fine-tuning pipeline, focusing on the effectiveness of the two regularization techniques across diverse domains.

\subsection{Experimental Settings}

\paragraph{Datasets and Evaluation Protocol.}
We evaluate our method on a broad spectrum of fine-grained visual domains to ensure its generality and robustness. Specifically, we use five open-set fine-grained image retrieval datasets as target domains for fine-tuning:

\begin{itemize}
    \item Stanford Online Products (SOP)~\cite{song2016deep} (product domain)
    \item InShop~\cite{liu2016deepfashion} (clothing domain)
    \item Stanford Cars196~\cite{krause20133d} (car domain)
    \item iNaturalist 2018~\cite{van2018inaturalist} (natural world domain)
    \item Food2k~\cite{min2023large} (food domain)
\end{itemize}
For consistency, we adopt the data splits from the Universal Embeddings Dataset (UnED)~\cite{ycc+23}. Each evaluation test set is partitioned into a query and index set for open-set nearest neighbor retrieval.

To rigorously evaluate the retention of generic, pretrained knowledge, we construct a benchmark where the model is fine-tuned on one target domain and evaluated across all other fine-grained domains, as well as on the coarser ImageNet domain, by evaluation on its unseen test set (using a leave-one-out retrieval setup). 
Additionally, to assess preservation of the model’s original vision-language alignment, we perform cross-modal retrieval (image-to-text and text-to-image) on the COCO~\cite{chen2015microsoft} and Flickr30k~\cite{young2014from} test sets, always keeping the text encoder frozen.

\paragraph{Metrics.}
For all image-only retrieval tasks, mAP@20 is reported. For text-image retrieval, Recall@1 (R@1) is reported, following established conventions. 
Textual embeddings are always produced by the original, frozen text encoder from the respective VLM backbone.
To summarize performance, we compute three aggregate metrics:
\begin{itemize}
    \item Image-Image In-Domain: mAP@20 on the fine-tuned domain’s test set (unseen classes, image-image retrieval).
    \item Image-Image Out-of-Domain (O-o-D) Average: average mAP@20 across the other fine-grained datasets and ImageNet (image-image retrieval).
    \item Text-Image Out-of-Domain (O-o-D) Average: average R@1 across all four text-image retrieval settings (2 datasets, both directions).
    \item In-Out Avg.: simple average of the In-Domain and overall Out-of-Domain scores (overall meaning average of Image-Image O-o-D Average and Text-Image O-o-D Average), providing a holistic measure of adaptation and generalization.
\end{itemize}

\paragraph{Pretrained Models.}
We evaluate our approach on two widely adopted Vision-Language Foundational Models, namely SigLIP~\cite{zhai2023sigmoid} and TIPS~\cite{maninis2025tips}. 
These backbones are chosen due to their strong and diverse performance across multimodal learning tasks, ensuring that our findings are not tied to a single pretraining.

\vspace{-10pt}

\paragraph{Baselines and Competitors.}

To contextualize the effectiveness of our method, we compare it against several baselines and competitive approaches:

\begin{itemize}
    \item Standard fine-tuning (``Standard''): Unconstrained adaptation using only the classification loss (Eq.~\ref{eqn:crossentropy}), as in~\cite{ycc+23}.
    \item L2-SP~\cite{xuhong2018explicit}: Uses only the parameter regularization loss ($\lameb=0$).
    \item WISE-FT~\cite{wortsman2022robust}: Parameter averaging between the pretrained and independently fine-tuned models.
    \item (Our) LDIFS variant~\cite{mukhoti2024ldifs}: Applies only the embedding regularization loss ($\lamtheta=0$), analogous to LDIFS (excluding intermediate layers for fairness).
\end{itemize}
We note that, for a better comparison, (Our) LDIFS variant selects the weight for the embedding regularization loss using our composite validation set, rather than solely in-domain validation as done in the original work. 
In such a case, it would fail to even enable the regularization loss, as a weight of 0 would maximize in-domain performance. 
This adjustment results in stronger out-of-domain results, as discussed in Section~\ref{valid}.
It is important to highlight that, to our knowledge, prior work has not systematically addressed the challenge of regularization weight selection~\cite{zheng2023preventing}, nor employed explicit out-of-domain validation in model selection~\cite{mukhoti2024ldifs}.

\vspace{-10pt}

\paragraph{Implementation details.}
All experiments use a batch size of 128, with learning rates of $10^{-5}$ for the backbone and $10^{-3}$ for the classifier prototypes. Prototypes are initialized with the embedding of the training sample whose representation is closest to the class mean embedding, computed from the pretrained network. 
An image resolution of $224 \times 224$ is used for both fine-tuning and evaluation, and adopt a ViT-Base backbone with a patch size of 16. 
The aforementioned hyperparameters follow prior work~\cite{ycc+23} and are not further tuned, as they are not central to our contributions.
For the TIPS visual encoder, we use the first of its two CLS tokens. While TIPS is pretrained with a patch size of 14, we use 16 for consistency across experiments; improved performance could be expected if using the default patch size that matches its pretraining setup.

The remaining hyperparameters are selected via grid search using our proposed validation protocol: $\lameb \in {10^2, 10^3, 10^4, 10^5}$ and $\lamtheta \in {10^3, 10^4, 10^5, 10^6}$. 
Our implementation builds on the Scenic framework~\cite{dga+21}, using JAX~\cite{jax2018github} and Flax~\cite{flax2020github}, and all experiments are conducted on Google Cloud TPU v2 and v3 hardware~\cite{norrie2021design}.

\begin{table*}[tbp]
    \centering
    \scalebox{0.96}{
    \begin{tabular}{c||c|c|c|c}
    \hline

    Fine-tuning & \multicolumn{2}{c|}{Image-Image (mAP@20)}  & \multicolumn{1}{c|}{Text-Img  (R@1)} \\ 
    Method     & In-Domain &  
    O-o-D Average & 
    O-o-D Average & In-Out Avg. \\ \hline
     
    Standard & \textbf{76.9 (+31.0)} &  44.9 (-6.0) & 55.7 (-12.8) & 63.6 (+10.8) \\

    L2-SP (parameters reg. only) & {74.7 (+28.8)} & {47.9 (-3.0)} & {60.8 (-7.6)} & {64.6 (+11.8)} \\
    
    {(Our) LDIFS - last-layer (in-domain embed. reg. only)} & {69.5 (+23.6)} & {49.5 (-1.4)} & {65.7 (-2.7)} & {61.6 {(+ 8.8)}}\\ 
    
    {(Our) LDIFS - last-layer (generic embed. reg. only)} & {75.9 (+30.0)} & {50.7 (-0.2)} & {66.6 (-1.8)} & {67.3 (+14.5)}\\
    
    WISE-FT (a=0.5) & 65.4 (+19.5) & \textbf{50.8 (-0.1)}  & 66.8 (-1.6) & 61.0 (+8.2) \\
    
    Ours & 75.8 (+29.9) & \textbf{50.8 (-0.1)} & \textbf{67.5 (-0.9)} & \textbf{67.5 (+14.7)}\\          
    \hline
    
    \end{tabular}
    }
    \caption{
    Comparison with different baselines, adapted from the previous work.
    All experiments start from SigLIP pretrained ViT-Base/16, fine-tuned on the SOP dataset.
    The performance difference ($\Delta$) to the pretrained model is reported next to the absolute performance.
    }
    \label{tab:competitors}

\vspace{-8pt}

\end{table*}

\subsection{Results.}
Unless otherwise noted, for each experiment the optimal pair of regularization weights $(\lameb,\lamtheta)$ is selected by grid search on the composite validation metric described in Section~\ref{valid}.

\vspace{-8pt}

\paragraph{Comparison against standard finetuning.}

Table~\ref{tab:main_table_all} summarizes the results of our method versus standard (unregularized) fine-tuning, across both SigLIP and TIPS pretrained models and all five fine-grained datasets. Our approach consistently achieves a strong balance between in-domain adaptation and out-of-domain generalization. Unlike standard fine-tuning, which quickly specializes but catastrophically forgets, our method preserves much of the model's original generalization and vision-language alignment, even though no text data or text encoder is used during fine-tuning. Notably, the optimal regularization weights differ by pretraining and target dataset, emphasizing the importance of principled hyperparameter selection, in contrast to earlier works that used fixed or uniform settings~\cite{zheng2023preventing, mukhoti2024ldifs}.

\vspace{-8pt}

\paragraph{Comparison against other baselines.}
Table~\ref{tab:competitors} provides a direct comparison with all baselines on SigLIP ViT-Base/16 fine-tuned on SOP. The results reveal distinct behaviors:
The unconstrained model achieves the best maximum in-domain performance but suffers from extensive out-of-domain knowledge degradation.
L2-SP (parameter regularization only) leads to a small drop in in-domain performance, but offers no significant retention in out-of-domain performance.
LDIFS (embedding regularization) with in-domain targets slightly improves out-of-domain scores but significantly restricts in-domain adaptation, since the embedding targets are drawn from the same data being specialized to.
Replacing in-domain with external images for embedding regularization (LDIFS with generic samples) improves both in-domain and out-of-domain performance by providing broader coverage of the pretrained embedding space.
WISE-FT (parameter averaging) retains good out-of-domain generalization, but sacrifices in-domain specialization.
Our method delivers the best trade-off, maintaining robust performance on the new domain without significant loss of generalization.

\paragraph{External dataset choice.} 
We further analyze the impact of the external dataset used for embedding regularization.
Using a SigLIP ViT-Base/16 fine-tuned on SOP, the regularization is performed on a subset of LAION (roughly the same size as ImageNet) as the external dataset instead, as shown in Table~\ref{tab:external_dataset}.
It should be noted that the validation set used in this experiment is the composed validation set of our method that uses ImageNet as the out-of-domain component. 
The reason is that LAION comes without class-level labels, so no retrieval metrics can be calculated on it.
The results support our choice of ImageNet: it provides comprehensive coverage of generic visual concepts, serving as an effective anchor for preserving pretrained capabilities during adaptation. 
Alternative datasets are possible, but ImageNet's accessibility and diversity make it particularly suitable for this purpose.

\begin{table}[h]
    \centering
    \scalebox{1.0}{
    \begin{tabular}{c||c}
    \hline    
    
    Generic dataset & In-Out Avg. \\ 
    \hline
    ImageNet & \textbf{67.5 (+14.7)}\\
    LAION & 65.5 (+12.7)\textbf{}\\          

    \hline
    
    \end{tabular}
    }
    \vspace{5pt} 
    \caption{
    Study for the external source of generic data used as targets for the pretrained embedding regularization.
    In this experiment, the full proposed method that uses both types of regularization is used.}
    \label{tab:external_dataset}
\end{table}

\vspace{5pt}
\vspace{-10pt}
\section{Conclusions}
In this work, a robust fine-tuning method for large Vision-and-Language Foundation Models that significantly improves performance on fine-grained visual domains while simultaneously preserving the rich, general-purpose knowledge acquired during large-scale multimodal pre-training was developed.
Two regularization techniques with different functionality and impact were used simultaneously. An analysis and an intuitive explanation of each of the regularizations were provided.
Additionally, methodological gaps present in prior work were addressed by introducing a rigorous validation strategy. 
This strategy explicitly evaluates both in-domain specialization and out-of-domain generalization performance, ensuring that hyperparameter tuning and early stopping decisions are principled and reproducible.

{
    \small
    \bibliographystyle{ieeenat_fullname}
    \bibliography{main}
}

\end{document}